\documentclass[letterpaper, 10 pt, conference]{ieeeconf} 
\IEEEoverridecommandlockouts    
\usepackage{graphics} 
\usepackage{epsfig} 
\usepackage{mathptmx} 
\usepackage{cite}
\usepackage{times} 
\usepackage{amsmath} 
\usepackage{amssymb}  
\usepackage{hyperref}

\usepackage{comment}
\usepackage{times}
\usepackage{multirow}
\usepackage{tabularx}
\usepackage{siunitx}
\usepackage{capt-of}
\usepackage{wrapfig}
\usepackage{url}
\usepackage{array}
\usepackage[table,xcdraw,dvipsnames]{xcolor}
\usepackage{eucal}

\newcommand{\mybullet}{\vspace{0.05cm}\noindent $\bullet$\ }
\newcommand{\mybulletend}{\vspace{0.05cm}}

\newcommand{\mysubsubsection}[1]{\vspace{0.1cm} \noindent {\bf #1}:}

\definecolor{darkgreen)}{rgb}{0.0, 0.5, 0.0}

\title{{\LARGE \bf Fusion-DHL: WiFi, IMU, and Floorplan Fusion for\\ Dense History of Locations in Indoor Environments}}

\author{Sachini Herath$^{1}$ \and Saghar Irandoust$^{1}$ \and Bowen Chen$^{1}$ \and  Yiming Qian$^{1}$ \and Pyojin Kim$^{2}$ \and Yasutaka Furukawa$^{1}$%
\thanks{$^1$ Simon Fraser University. {\{sherath, sirandou, bowenc, yimingq, furukawa\}@sfu.ca}}
\thanks{$^2$ Sookmyung Women's University. {pjinkim@sookmyung.ac.kr}}
\thanks{This research is partially supported by NSERC Discovery Grants, NSERC Discovery Grants Accelerator Supplements, and DND/NSERC Discovery Grant Supplement.}
}

\let\oldtwocolumn\twocolumn
\renewcommand\twocolumn[1][]{%
    \oldtwocolumn[{#1}{
    \begin{center}
    \includegraphics[width=\textwidth]{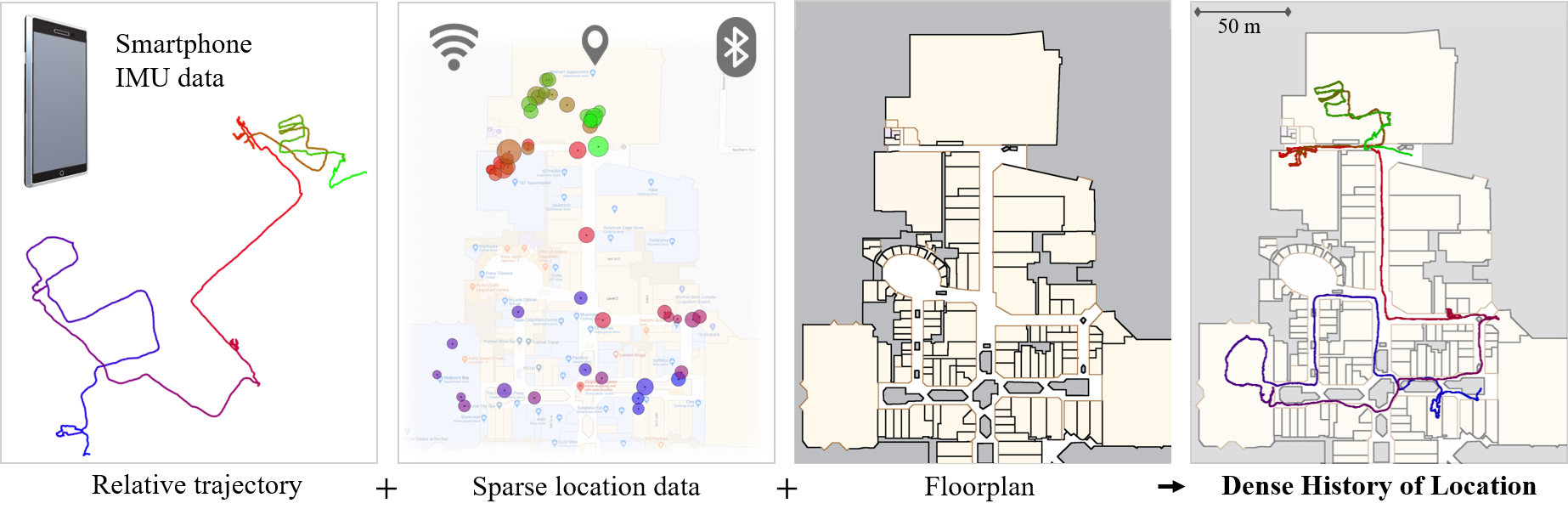}
 \captionof{figure}{The paper proposes a novel multi-modal sensor fusion algorithm that fuses 1) a relative motion trajectory by inertial navigation algorithm based on IMU sensor data; 2) sparse location data by geo-localization system based on WiFi; and 3) a floorplan image. The system works with any modern smartphone and is able to produce dense history of location (DHL) with minimal additional energy consumption on the smartphone. (The color of the trajectory and points, blue→red→green, encodes time.)}
 \label{fig:teaser}
        \end{center}
    }]
}

\begin{document}

\maketitle
\begin{abstract}
The paper proposes a multi-modal sensor fusion algorithm that fuses WiFi, IMU, and floorplan information to infer an accurate and dense location history in indoor environments.
The algorithm uses 1) an inertial navigation algorithm to estimate a relative motion trajectory from IMU sensor data; 2) a WiFi-based localization API in industry to obtain positional constraints and geo-localize the trajectory; and 3) a convolutional neural network to refine the location history to be consistent with the floorplan.
We have developed a data acquisition app to build a new dataset with WiFi, IMU, and floorplan data with ground-truth positions at 4 university buildings and 3 shopping malls.
Our qualitative and quantitative evaluations demonstrate that the proposed system is able to produce twice as accurate and a few orders of magnitude denser location history than the current standard, while requiring minimal additional energy consumption.
We will publicly share our code and models.
(\url{https://github.com/Sachini/Fusion-DHL})
\end{abstract}

\begin{figure*}[t]
    \centering
    \includegraphics[width=\textwidth]{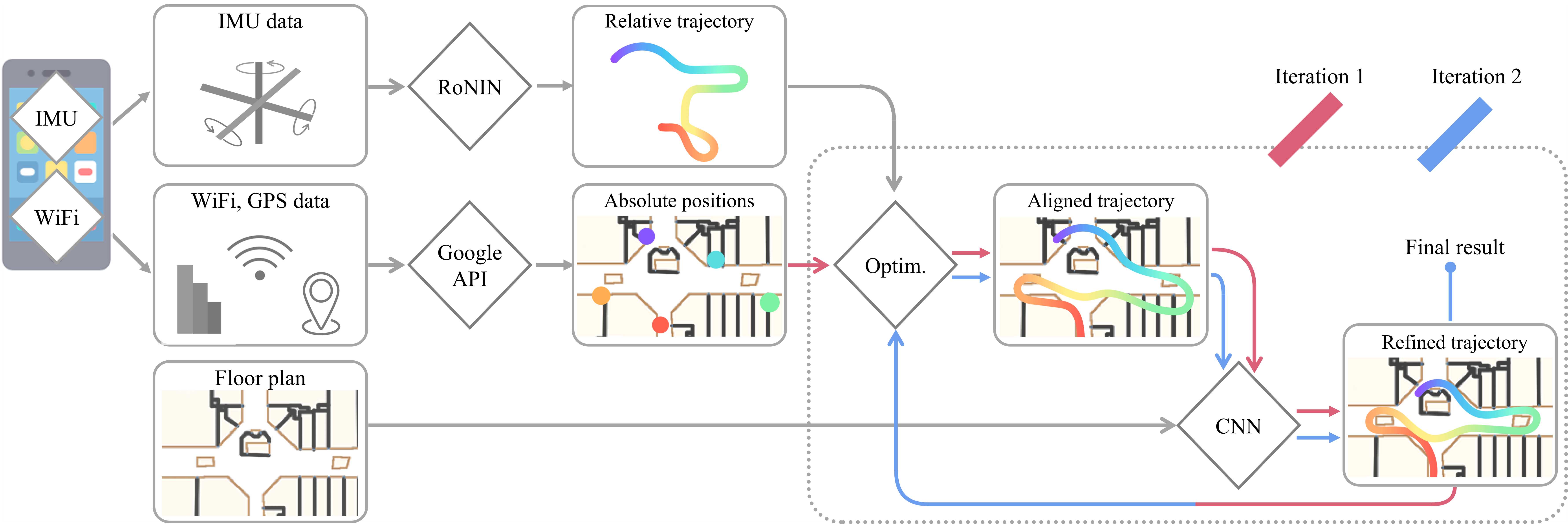}
    \caption{System pipeline: We use a data-driven inertial algorithm to estimate relative motion trajectories from IMU sensor data. We geo-localize the trajectory with a floorplan by solving an optimization problem subject to the constraints from the state-of-the-art Indoor GPS system (i.e., Google Fused Location Provider). We use a convolutional neural network to refine the trajectory (time coded by color, blue→red) to be consistent with a floorplan. The system repeats the optimization and the CNN application once more to further improve the accuracy of the location history.}
    \label{fig:system}
\end{figure*}

\section{Introduction}
Indoor positioning has been a long-awaited technology, enabling mobile location-aware services, augmented reality applications, and location history collection. The location history is, in particular, getting more attentions for contact-tracing upon COVID pandemic. The location history further enables space utilization studies for indoor space planning or traffic flow analysis for transportation optimization.

Contemporary indoor positioning systems utilize multi-modal sensor data such as global positioning system (GPS), inertial measurement unit (IMU), and building dependent infrastructures such as WiFi, Bluetooth or magnetic fingerprints. WiFi map is the most informative, which holds pairs of access-point IDs (i.e., MAC address) and their global positions (i.e., latitude/longitude degrees). The quality of WiFi-maps directly influence the accuracy of indoor positioning systems, where the community~\cite{wigle} and digital mapping companies (e.g., Google Maps and Apple Maps) are competing to build better WiFi maps.

This paper takes the indoor positioning technology to the next level, specifically focusing on the location history estimation as an offline task. Our idea is the novel fusion of WiFi, IMU, and floorplan data. Concretely, we use 1) a state-of-the-art inertial navigation algorithm (RoNIN~\cite{ronin}) to estimate a relative motion trajectory from IMU; 2) Google Fused Location Provider API, which mostly relies on WiFi, to obtain sparse global positions and align the inertial trajectory to a map; and 3) a convolutional neural network to refine the location history to be consistent with a floorplan.
Our system works with any modern smartphone, while requiring minimal extra energy consumption on the device, where energy-efficient IMU is the only additional data acquisition.

We have built a new benchmark with WiFi, IMU, and floorplan data at 4 university buildings (training sites) and 3 shopping malls (testing sites), spanning 15.2 hours and 42.3 kilometers. Dense ground-truth positions are obtained by manually geo-localizing relative motion trajectories from a visual inertial SLAM algorithm. For long sequences through shopping malls in the wild (i.e., as long as 45 minutes), the ground-truth positions are obtained for a sparse set of points by manual specifications.Our qualitative and quantitative evaluations demonstrate that the proposed system is able to produce twice as accurate (i.e., RMSE roughly 5m instead of 12m) and a few orders of magnitude denser (i.e., 50Hz instead of 1/60Hz) location history than the current state-of-the-art system.

\section{Related Work}
We discuss related works in three areas: inertial navigation, indoor localization via wireless networks, and indoor localization via multi-modal sensor fusion.

\mysubsubsection{Inertial navigation} 
IMU is energy efficient and works anytime anywhere.
Inertial navigation has been a dream technology for many years.
Classical algorithms (also known as Pedestrian Dead Reckoning) are either physics based (e.g., IMU double integration), or heuristic based (e.g., zero speed update~\cite{jimenez2009zupt} or step counting~\cite{step_counting}).
With the surge of deep learning, data-driven approaches~\cite{ronin, yan2018ridi, chen2018oxiod} have made a breakthrough. A state-of-the-art system
achieves the positional accuracy of 5 meters per minute under natural activities in the wild~\cite{ronin}.
The major error source is the rotational drift due to gyroscope bias.

\mysubsubsection{Indoor localization via wireless networks} 
Wireless communication networks (e.g., WiFi~\cite{yang2015wifi} or Bluetooth~\cite{chen2016ibeacon}) are the most reliable information for indoor localization. A system requires
either 1) an access-point database (i.e., access-point IDs and their geo-coordinates), which allows trilateration or weighted averaging of access-point coordinates~\cite{wigle}; or 2) a fingerprint database (i.e., received signal strength indicators and geo-coordinates at their observations), which  allows fingerprint-based geo-coordinate look-ups~\cite{bahl2000radar,wu2012will}.
Google Fused Location Provider API (FLP) \cite{flp} is a widely used system, which we believe is mostly WiFi-based. However, the system is energy-draining for a dense location history and often exhibits positional errors of more than 10 meters. 

\mysubsubsection{Indoor localization via Multi-modal sensor fusion} 
Modern smartphones are equipped with a suite of sensors. 
WiFi-SLAM \cite{huang2011wifigraphslam, mirowski2013signalslam, ferris2007wifi, kotaru2015spotfi} minimizes accumulation errors in inertial navigation by WiFi-based loop-closing constraints.
Magnetic field distortions are specific to building infrastructures and provide additional cues for the fingerprint-based system~\cite{solin2018magnet}.
Map-Matching algorithms align motion trajectories with a floorplan via 
conditional random fields~\cite{xiao2014lightweight}, hidden Markov models~\cite{thiagarajan2009vtrack, newson2009hidden}, particle filtering~\cite{davidson2010application}, or dynamic programming~\cite{xiao2014lightweight}.
The paper proposes a novel fusion of WiFi, IMU, and floorplan data by a combination of optimization and a convolutional neural network (CNN).

\section{System: Dense History of Location Inference}
\label{sec:system}
Our system, coined DHL, works with any modern smartphone, utilizing an IMU and a WiFi receiver (See Fig.~\ref{fig:system}). The input to the system is IMU sensor data, global positions by a geo-localization API such as Google FLP at 1/60Hz, and a geo-localized floorplan.
This section provides the system overview, which consists of the following steps.

\vspace{0.2cm}

\noindent
$\bullet$
A data-driven inertial navigation algorithm (RoNIN~\cite{ronin}) estimates a relative motion trajectory at 50Hz from IMU.

\noindent
$\bullet$
A non-linear least squares optimization is used to geo-localize the trajectory subject to positional constraints by FLP and regularization terms penalizing the deviations from the original trajectory with a sensor-centric error model.
Besides the geo-localization, this alignment removes scale and/or rotational drifts inherent to the inertial navigation.

\noindent
$\bullet$
A convolutional neural network (CNN) is trained to estimate a correction to the aligned trajectory for being consistent with a floorplan. 

\noindent
$\bullet$
Lastly, the system runs the optimization and the CNN steps once more, while changing the positional constraints from FLP to the current location history estimation, subsampled once every 200 frames.

\section{Dense Location History Dataset}
\label{sec:dataset}
The first contribution of the paper is a novel multi-modal sensor dataset with ground-truth positions.
Our dataset contains 9.6 hours (26.1 km) of motion data over 4 university buildings (6 floorplans) and 5.7 hours (16.2 km) of motion data over 3 shopping malls (6 floorplans).
Shopping malls are the indoor environments with immense business opportunities and our interests, which are kept for testing.
University data are used for training a CNN module, where a single trajectory spans 5-10 minutes. Mall data are much longer (i.e., 15-45 minutes) and hence more challenging.

{\bf IMU} data consists of accelerations (200Hz), angular velocities (200Hz), and device orientations (100Hz).
A subject walks on a single floor of a building, while carrying a smartphone naturally by a hand or in a pocket (either Samsung Galaxy S9 or Google Pixel 2 XL).
We have not performed gyro-calibration (e.g., placing on a table for 10 seconds) in advance
to simulate the real-world scenarios, which causes severe ``bending'' due to rotational accumulation errors.

{\bf WiFi}-based geo-positions are collected by the Google Fused Location Provider (FLP). 
We call the FLP API at 1Hz to obtain geo-positions as much as possible for evaluation. This rate is impractical for real-world applications as the API drains battery. We subsample FLP data at a realistic rate (1/60Hz) as an input to a system.

{\bf Floorplan} images are created based on blueprints for university buildings and Google maps for shopping malls (See Fig.~\ref{fig:teaser}).
Corridors, rooms, and unwalkable areas are in white, yellow, and grey, respectively. Open boundaries (e.g., walls with doors and storefronts) are in brown. The other walls are in black. University and shopping-mall floorplans are scaled so that 5 and 2.5 pixels correspond to a meter, respectively.
We align the floorplans with the global coordinates by manually specifying correspondences between the floorplans and the Google Maps.

{\bf Ground-truth} (GT) positions for shopping malls are obtained as either a sparse set of positions from real-time user clicks or dense trajectories from visual inertial SLAM.
First, a set of ground-truth positions 
are obtained through an Android app where a subject clicks a floorplan on a smartphone real-time during the data acquisition. At total, 206 GT positions are collected for 5.0 hours of data from the three buildings, where a single trajectory spans 27.3 minutes on the average. Second, in one of the floorplans, we use visual inertial SLAM of Asus Zenfone (Google Tango phone) to collect relative trajectories, 
which are geo-localized by a rigid transformation with manual offline specifications to generate dense GT data.
Due to the memory limitation on the visual SLAM system, the trajectories with dense GT positions are much shorter and hence less challenging, where a single trajectory spans 4.2 minutes on the average.

University data (9.6 hours) are used only for training a CNN module, where dense GT positions are obtained by geo-localizing inertial navigation trajectories with real-time user clicks by an optimization algorithm in our system (See Sect.\ref{sec:optimizer} for details).
Note that GT positions may not be accurate for frames away from the user clicks. However, the university data are used only for training, and do not affect our evaluation. In practice, this training data allows the CNN model to generalize reasonable well to shopping malls.

\begin{figure*}[tb]
    \centering
    \includegraphics[width=\linewidth]{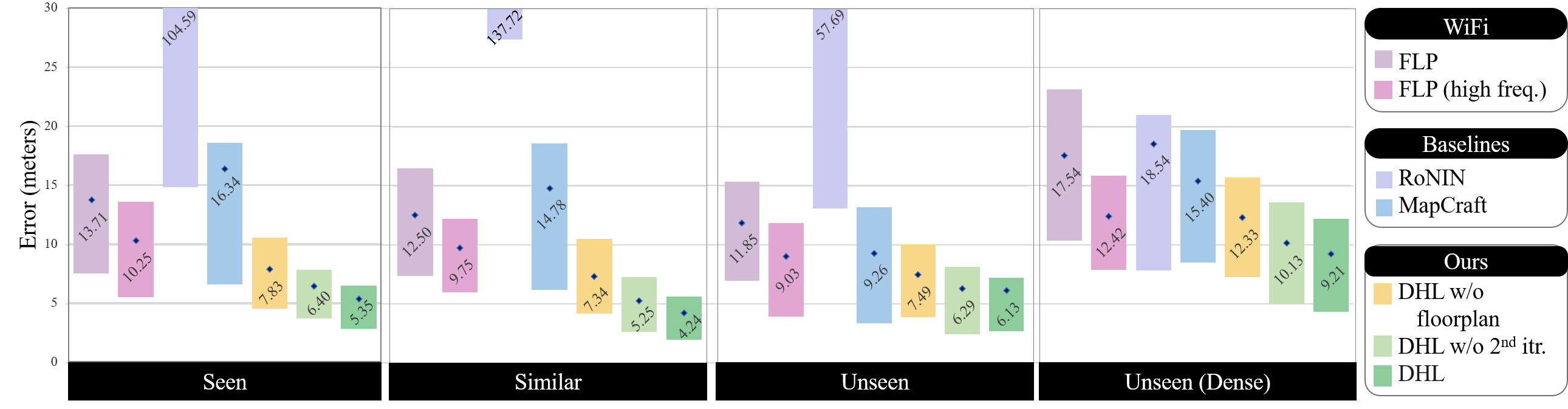}
    \caption{Quantitative evaluation: Localization accuracy for 4 groups of test data (data groups described in \ref{sec:quant}). We compare the proposed approach (DHL) against four competing methods and two variants of our system. Each bar shows the mean and the first/third quartiles of the localization errors.
    }
    \label{fig:quantitative}
\end{figure*}

\section{Algorithm: WiFi, IMU, and Floorplan Fusion} \label{sec:algorithm}

The section explains the details of the non-linear least squares optimization, which fuses time-synchronized WiFi and IMU data, and the convolutional neural network, which fuses floorplan information for refinement.

\subsection{IMU and WiFi Fusion by Optimization} 
\label{sec:optimizer}

A relative motion trajectory by RoNIN is a sequence of speed $s_f$ and motion direction angle $\theta_f$ at every frame $f$. The position 
of the trajectory at frame $f$ is calculated as
$\sum_{i=1}^f s_i \left[\cos{\theta_i}, \sin{\theta_i}\right]^T$,
assuming that motions are 2D on a single floor and start from the origin. The units of the positional and angular quantities are meters and radians in our formulation, respectively.

Inertial navigation exhibits two types of errors: 1) accumulation errors in the device orientation due to gyroscope bias; and 2) scale inference errors (i.e., the walking speed varies per person). We 
solve for the scale correction $\Delta s_f$, the angle correction $\Delta \theta_f$, and the positional displacement at the first frame $\Delta \overrightarrow{P_0}$, where the corrected position is calculated as
\begin{gather*}
\overrightarrow{P_f} = \Delta \overrightarrow{P_0} + \sum_{i=1}^f s_i \cdot \Delta s_i \left[\cos{\left(\theta_i + \Delta \theta_i\right)}, \sin{\left(\theta_i + \Delta \theta_i\right)}\right]^T.
\end{gather*}
Note that the scale (resp. angle) correction terms should be smooth, and we define a correction term every 100 (resp. 20) seconds, while using linear interpolation to derive the amount of correction at an arbitrary frame~\cite{yan2018ridi}. In reality, angle corrections should be on the device axes, but we make a simplified assumption and directly model corrections in the heading angle around the gravitational axis.

Concretely, we formulate a non-linear least squares optimization problem subject to the positional constraints from FLP and the regularization constraints $(w_1 = 10, w_2 = 200)$:
\begin{gather*}
     \sum_{f\in\mathcal{F}_{FLP}} max(\|\overrightarrow{P_f} - \overrightarrow{P^{FLP}_f}\|^2-r^{FLP}_f, 0)+ w_1
      E^{REG}_1\left( \{\Delta s_f\}\right) \\+ w_2 E^{REG}_2\left(\{\Delta \theta_f\}\right)
\end{gather*}
The first term is the sum of squared distances between the corrected position ($\overrightarrow{P_f}$) and the geo-position by FLP. We only penalize if the error is larger than accuracy reported by FLP, $r^{FLP}_f$.
$\mathcal{F}_{FLP}$ denotes the set of frames with FLP constraints.
The second term is the regularization over the scale corrections, where $\Delta s_f$ should be close to 1.0. The term is $\max(\Delta s_f, \frac{1}{\Delta s_f})^2$, summed over all the variables. We also set $s_f \ge 0$ as constraints. 
The last term is the regularization over the angle corrections.
We simply use squared magnitudes of the first order and the second order derivatives via finite differences over the variables. The second order magnitude is further amplified by a factor of 1.5.

Before the optimization, we solve for a rigid transformation by linear least squares 
that aligns the FLP positions and the corresponding frames in the inertial navigation trajectory. The rigid transformation is used to initialize the angle correction and the positional displacement variables, while keeping the scale correction variables to be $1.0$.

\subsection{Floorplan fusion by CNN}
\label{sec:DNN}
FLP positional errors are often 10 meters or more. We refine the aligned trajectory based on the floorplan via a convolutional neural network (CNN). The section explains the data representation, the network architecture, and the pre-processing steps for training and testing.

\mysubsubsection{Data representation}
We represent each trajectory segment as an image by scatter-plotting the curve in the black background with disks of 3 pixel-radius and with a rainbow colormap representing time (see Fig. 1). Being concatenated with a RGB floorplan image, the input is a 6-channel image of size $250 \times 250$.
The output is a ``correction flow'', which is a 2D positional displacement vector, only defined for each pixel along the input trajectory segment (other pixel values are disregarded when computing loss and generating the corrected trajectory). The correction flow is represented as a 2-channel image of size $250\times 250$, where the numeric values are in the unit of a pixel.

\mysubsubsection{Network architecture}
We use U-Net~\cite{unet}, in particular, an implementation for brain image segmentation~\cite{unet_code} with one modification. We added a bilinear layer to our input image, which increase the number of channels from 6 to 18.
We use the Smooth L1 Loss with a default PyTorch implementation at every pixel along the input segment, where no loss is defined at the remaining pixels.

\mysubsubsection{Training data pre-processing} Our training data come from university buildings, while testing data come from shopping malls. Both have similar error characteristics on the inertial trajectories, but FLP behaves differently.
We use university data to generate training samples that generalize well to shopping mall data by simulating FLP noise.~\footnotemark

To study the generalization capability of the system over different building types (i.e., from university buildings to shopping malls), we also generate synthetic training data by drawing trajectories on one mall floorplan by a Python GUI.
Specifically, we use cubic splines to generate a continuous and constant-velocity trajectory, which is treated as the ground-truth. The input ``corrupted'' trajectories are generated in the same way as before, after adding synthetic bias/scale errors. In total, we have generated 4,556 training samples from university buildings and 825 samples from shopping malls (synthetic).

~\footnotetext{
First, we randomly select a reference frame from the first 85\% of an inertial navigation trajectory from a university building, then geo-localize the trajectory by aligning the position and the motion direction at the reference frame with the ground-truth. We follow the trajectory from the reference and crop when it reaches the last frame or within 5 pixels from the border of the $250\times 250$ pixel square centered at the reference frame.
We simulate FLP positional error by a 2D Gaussian function with a standard deviation set to 25 pixels (= 5 meters), because FLP at shopping malls exhibits roughly 10 meters positional errors. More concretely, we take the trajectory segment, perturb the first and the last positions of the cropped segment by the Gaussian noise, and solve for a scaled rigid transformation based on the two points to warp the trajectory. We repeat the process and generate twenty samples from each inertial navigation trajectory. Lastly, we apply random horizontal flipping and a random rotation.
}

\mysubsubsection{Testing data pre-processing} An inertial trajectory can be as long as 45 minutes. We divide into overlapping shorter segments by 1) either taking the first 4 minutes or a segment whose axis aligned bounding box does not fit inside $250\times 250$ pixel square of the floorplan, whichever comes first; and 2) repeat the process after removing the three quarters of the segment to create overlaps. For each segment, we crop a $250\times 250$ pixel region of a floorplan so that the bounding box of the trajectory is centered. 
Pixels in the overlapping regions have multiple predictions from the CNN model, where we take their weighted average. Suppose a trajectory segment spans a time interval of $\left[0, T\right]$. The weight of a pixel is given by a normal distribution $\mathcal{N}(T/2, T/4)$ based on its time. 
\begin{figure*}[tb]
    \centering
    \includegraphics[width=\linewidth]{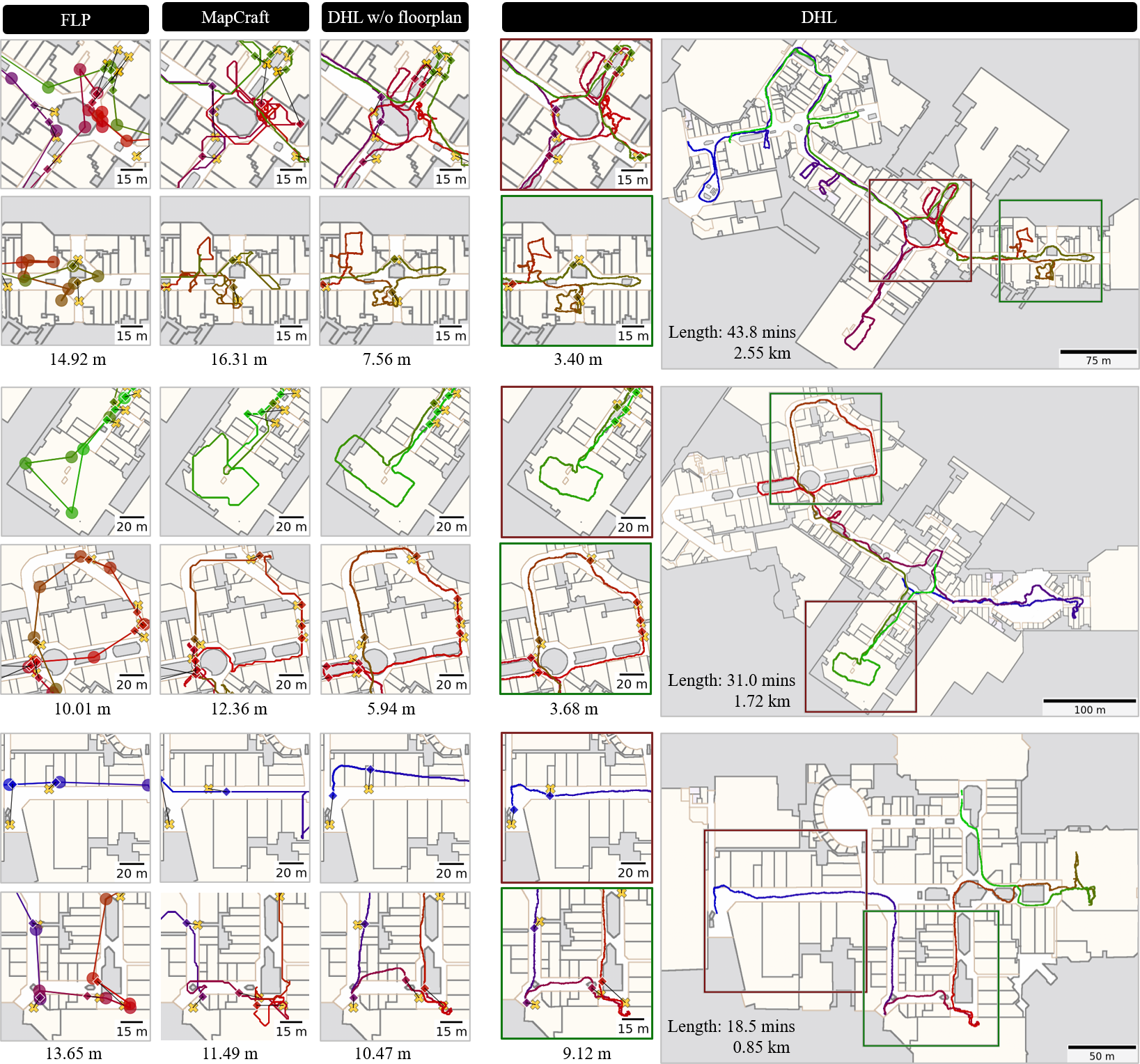}
    \caption{Qualitative evaluation: Visualization of inferred dense location history. DHL is the proposed system at the right.
    We compare against three other methods in close-ups.
    The color of the trajectory (blue→red→green) encodes time.
    For FLP, the trajectory estimation is drawn as a piecewise linear curve, connecting the FLP positions, which are highlighted as circles.
    For each figure, the ground-truth positions are shown as yellow 'x's, and the corresponding inferred positions are highlighted with square markers. The average localization error is shown by the number at the bottom.}
    \label{fig:qualitative}
    \vspace{-1mm}
\end{figure*}

\begin{figure*}[t]
    \centering
    \includegraphics[width=\linewidth]{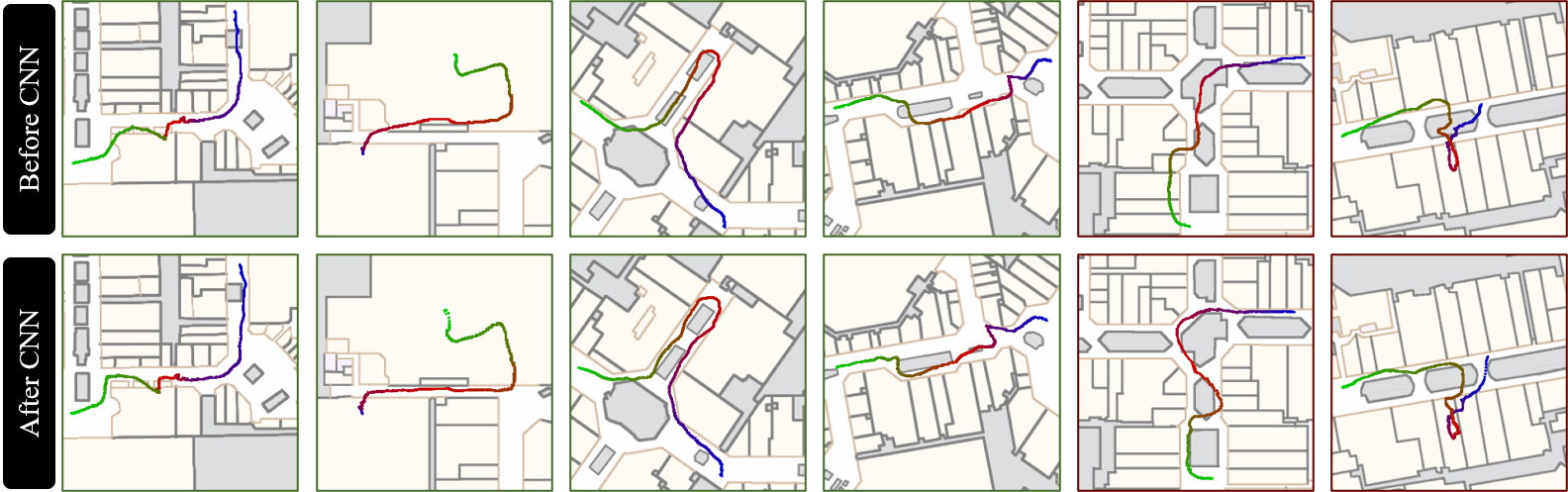}
    \caption{Floorplan fusion by CNN. The top and the bottom shows the trajectories (time coded by color, blue→red→green) before and after the CNN application, respectively.
    CNN is effective in avoiding collisions with walls or inaccessible areas (first 4 columns). The last two columns show failure cases due to large initial alignment errors.}
    \label{fig:cnn_result}
    \vspace{-4mm}
\end{figure*}

\section{Experimental results}
\label{section:results}
We have implemented the proposed system in Python and Pytorch, and used a workstation with a GeForce RTX 2080 Ti GPU.
For CNN training, we have used Adam optimizer~\cite{adam}. The learning rate has been initialized to $0.001$ and reduced by a factor of 10 after every 10 epochs when the loss does not decrease.
The total training time is roughly 4 hours.
For RoNIN software, we have downloaded the trained model from the official website~\cite{ronin}. Scipy least-squares function has been used for the optimization~\cite{2020SciPy-NMeth,optimizationLM}.

\subsection{Quantitative Evaluation}
\label{sec:quant}
We have compared the localization accuracy against the four competing methods (See Fig.~\ref{fig:quantitative}):

\mybullet {\bf RoNIN} is the inertial navigation trajectory, geo-localized based on the first minute of FLP constraints.

\mybullet {\bf FLP-high-freq} is a piece-wise linear trajectory, connecting the FLP locations that are queried at every second.

\mybullet {\bf FLP} is the same as FLP-high-freq, except that FLP is queried once every minute for energy efficiency.

\mybullet {\bf Mapcraft} is a state-of-the-art map-matching system~\cite{xiao2014lightweight}.~\footnotemark
\mybulletend
\vspace{0.05cm}

\footnotetext{
The code is not publicly available and we implemented locally with a few modification to improve accuracy: 1) Using
the RoNIN result as the velocity input; 2) Increasing the search neighborhood by a factor of 1.5 to handle scale inaccuracies in inertial navigation; and 3) Using FLP positions as a feature.
MapCraft is memory intensive and does not scale to large floorplans.
We developed a heuristic to crop the floorplan based on the FLP positions.
The edge of a graph is set to 1 meter.
}

Our test-sites are shopping malls, where the dataset contains 3 buildings with 2 floorplan images each, where we create four testing setups for assessing the generalization capability over different building types. Let us denote the six floorplan images as (A1, A2), (B1, B2), and (C1, C2), where a letter denotes a building. Suppose synthetic training data was generated on A1.

\mybullet{\bf Seen} group is (A1: 1.7 hours). 

\mybullet{\bf Similar} group is (A2: 1.4 hours), as the overall building structure is similar (i.e., 1st floor vs. 2nd floor).

\mybullet{\bf Unseen} group is (B1, B2, C1, C2: 2.0 hours), where buildings are different. 

\mybullet{\bf Unseen-dense} is (B1, B2: 0.7 hours), one unseen building with dense ground-truth by visual inertial SLAM.

\mybulletend

Fig. \ref{fig:quantitative} shows the mean and the first/third quartiles of the localization errors in meters. DHL (w/o Floorplan) is the result of our system after the WiFi/IMU fusion. In comparison to FLP, whose primary information source is WiFi, the result demonstrates the effectiveness of integrating inertial navigation, where the localization accuracy improves by roughly 5 meters (40\%). DHL (w/o 2nd itr.) is our system with one iteration of optimization and CNN, which is already better than all the competing methods. DHL shows that an extra iteration further improves the accuracy for about a meter in many cases.

FLP-high-freq indicates the limit of the current industrial state-of-the-art when one affords to call the API every second, which drains battery. The method improves the accuracy over FLP (API call every minute) by roughly 3 meters, but is inferior to DHL by 3 to 5 meters (DHL improves by 25\% to 50\%), demonstrating the effectiveness of our WiFi/IMU/floorplan fusion approach. MapCraft suffers from
discretization errors and the lack of holistic image understanding capabilities like CNN, where the result is inferior to even FLP-high-freq. 
RoNIN is the raw inertial navigation trajectory, which suffers from severe bending due to the accumulation errors in the device orientation, where the localization errors are worse than 50 to 100 meters.

Unseen-dense evaluation also supports the claims where the fusion of WiFi, IMU, and floorplan consistently improves the localization accuracy, where our method (DHL) is the most accurate.
Note that FLP accuracy is worse in the Unseen-dense setup especially for FLP, probably because the 3D tracking phone uses Android 8 while others use Android 10. Furthermore, the data was collected about a year before for Unseen-dense.

Lastly, we have evaluated the contribution of the synthetic training data. The average localization error of DHL increases from 5.57 meters to 6.85 meters on the average over the three main setups (Seen, Similar, and Unseen) when the synthetic data is not used.

\subsection{Qualitative Evaluation}
Figure~\ref{fig:qualitative} demonstrates how the multi-modal sensor fusion improves the results qualitatively. DHL w/o floorplan shows that WiFi (i.e., FLP positions) successfully removes the severe bending and roughly geo-localizes the inertial navigation trajectory.
DHL in comparison to DHL w/o floorplan shows that the floorplan information further refines the trajectory to be more accurate. Figure~\ref{fig:cnn_result} directly compares the results before and after the CNN-based floorplan fusion. Our CNN effectively avoids walking through walls/escalators and keeps the trajectory within the walk-able areas. Our CNN is sometimes able to correct the trajectory over 15 pixels (6 meters).
High frequency local motion detail is another difference in our results in contrast to FLP and MapCraft. As demonstrated in the supplementary video, together with the body orientation estimation by RoNIN system, DHL could potentially allow us to calculate the wall that receives the most human gazes and hence is the best place to put advertisement. Please refer to the supplementary material for more renderings.
\section{Conclusion}
The  paper  proposes  a  multi-modal  sensor  fusion algorithm that fuses WiFi, IMU, and floorplan information to infer dense  location  history, together with the dataset with the ground-truth. In contrast to the current industrial state-of-the-art (i.e., Google Fused Location Provider), our result is a few orders of magnitude denser and twice as accurate.
The current system is designed as an offline process generating location history as opposed to real-time Indoor GPS system. 
Our future work is the development of real-time Indoor GPS system via multi-modal sensor fusion.
\clearpage

\bibliographystyle{IEEEtran}
\bibliography{bibtex/bib/IEEEabrv,bibtex/bib/egbib}

\end{document}